\documentclass[conference]{IEEEtran}
\IEEEoverridecommandlockouts
\usepackage{cite}
\usepackage{amsmath,amssymb,amsfonts}
\usepackage{algorithmic}
\usepackage{graphicx}
\usepackage{textcomp}
\usepackage{xcolor}
\usepackage{threeparttable}
\usepackage{booktabs} 
\usepackage{graphicx} 
\usepackage{subcaption}
\usepackage{url}

\def\BibTeX{{\rm B\kern-.05em{\sc i\kern-.025em b}\kern-.08em
    T\kern-.1667em\lower.7ex\hbox{E}\kern-.125emX}}
\begin{document}

\title{Self-Reinforced Graph Contrastive Learning
}

\author{
    \IEEEauthorblockN{Chou-Ying Hsieh\IEEEauthorrefmark{1}, Chun-Fu Jang\IEEEauthorrefmark{1}, Cheng-En Hsieh\IEEEauthorrefmark{1}, Qian-Hui Chen\IEEEauthorrefmark{1}, Sy-Yen Kuo\IEEEauthorrefmark{1}\IEEEauthorrefmark{2}}
    \IEEEauthorblockA{\IEEEauthorrefmark{1}Department of Electrical Engineering, National Taiwan University 
    \\\{f07921043, r12921060, r12943101, r13921039, sykuo\}@ntu.edu.tw}
    \IEEEauthorblockA{\IEEEauthorrefmark{2}Department of Computer Science and Information Engineering, Chang Gung University
    \\kuosy@cgu.edu.tw}
}


\maketitle

\begin{abstract}
Graphs serve as versatile data structures in numerous real-world domains—including social networks, molecular biology, and knowledge graphs—by capturing intricate relational information among entities. Among graph-based learning techniques, Graph Contrastive Learning (GCL) has gained significant attention for its ability to derive robust, self-supervised graph representations through the contrasting of positive and negative sample pairs. However, a critical challenge lies in ensuring high-quality positive pairs so that the intrinsic semantic and structural properties of the original graph are preserved rather than distorted. To address this issue, we propose SRGCL (Self-Reinforced Graph Contrastive Learning), a novel framework that leverages the model’s own encoder to dynamically evaluate and select high-quality positive pairs. We designed an unified positive pair generator employing multiple augmentation strategies, and a selector guided by the manifold hypothesis to maintain the underlying geometry of the latent space. By adopting a probabilistic mechanism for selecting positive pairs, SRGCL iteratively refines its assessment of pair quality as the encoder’s representational power improves. Extensive experiments on diverse graph-level classification tasks demonstrate that SRGCL, as a plug-in module, consistently outperforms state-of-the-art GCL methods, underscoring its adaptability and efficacy across various domains.
\end{abstract}

\begin{IEEEkeywords}
graph neural network, graph contrastive learning, expectation-maximization algorithm, manifold hypothesis, self-reinforcement
\end{IEEEkeywords}

\section{Introduction}

Graphs, as powerful and flexible data structures, have become a cornerstone in numerous real-world applications, ranging from social networks \cite{myers2014information} and biochemical molecule modeling \cite{borgwardt2005protein} to knowledge graphs \cite{hogan2021knowledge} and recommendation systems \cite{fan2019graph}. Their ability to capture complex relational information among entities grants a high level of generalizability, enabling practitioners to tackle diverse tasks under a unified framework. Consequently, graph-based learning has gained remarkable attention for its capacity to model and analyze data where relationships play a critical role. 

\begin{figure}
    \centering
    \includegraphics[width=0.85\linewidth]{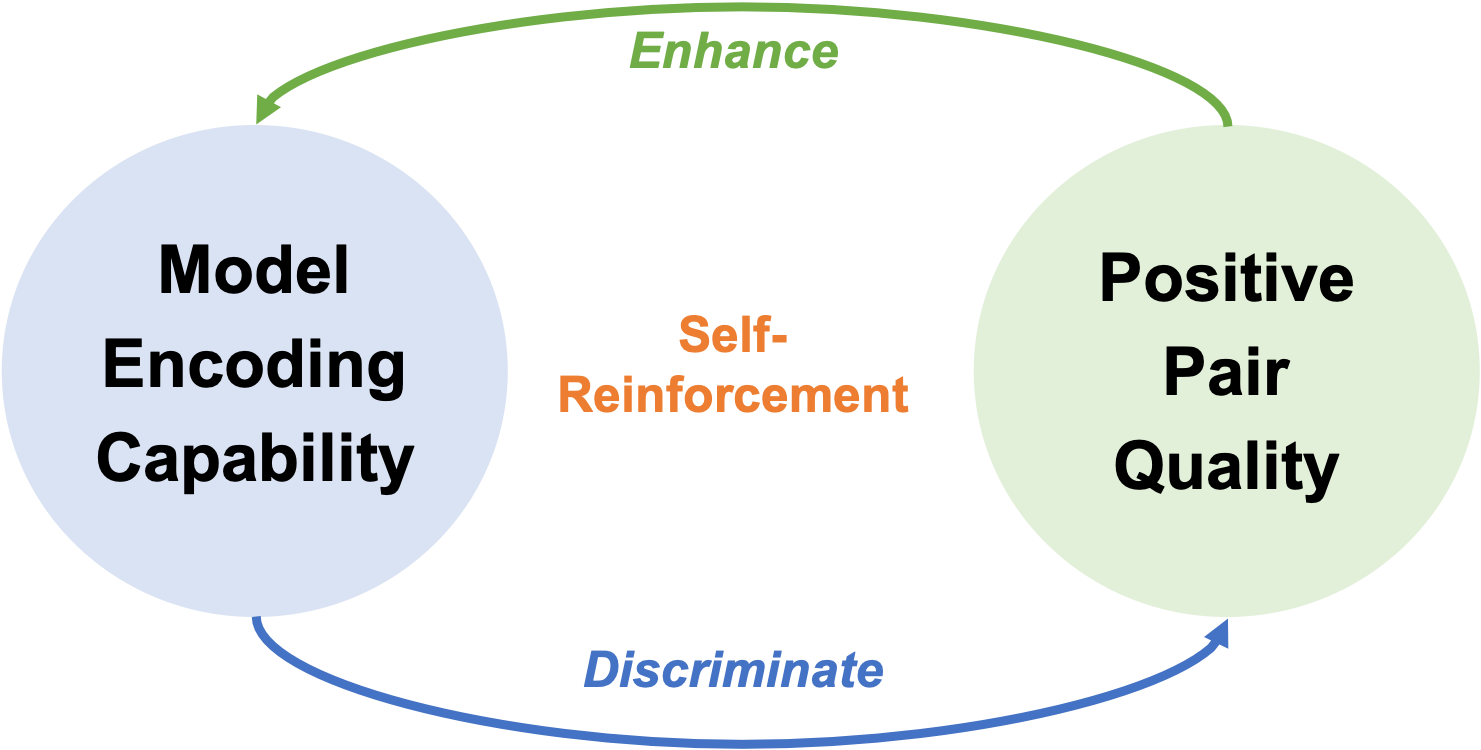}
    \caption{The core idea of SRGCL is that high-quality positive pairs enhance the model’s capability, enabling it to more effectively discriminate the quality of positive pairs.}
    \label{fig:main_idea}
\end{figure}

In Graph Contrastive Learning (GCL) \cite{you2020graph,hassani2020contrastive,sun2019infograph,xia2022simgrace,you2021graph,suresh2021adversarial,yin2022autogcl}, a \emph{positive pair} comprises two correlated views of the same underlying graph. We regard a pair as \emph{high-quality} when (i) both views conserve critical topological or attribute-based substructures and (ii) the two graphs necessarily remain in the same semantic class as the anchor. Conventional random augmentations (e.g., node dropping or edge perturbation \cite{you2020graph}) easily violate these conditions, producing \emph{false positives} that mislead the contrastive objective. A typical failure case is the MUTAG dataset \cite{debnath1991structure}, where randomly deleting bonds can break aromatic rings and invert mutagenicity labels, degrading representation quality. Empirically, substituting the two augmented views in GraphCL with two ground-truth graphs sharing the same label boosts accuracy from $87.59\!\pm\!0.78\%$ to $89.02\!\pm\!1.29\%$, and similar gains appear across other benchmarks, underscoring the importance of high-quality positives.

Despite recent progress, existing GCL frameworks offer limited mechanisms to guarantee such quality, especially as graph size and heterogeneity increase. Unreliable positives hamper the encoder from learning discriminative features and ultimately yield sub-optimal graph-level embeddings.

To overcome this limitation, we introduce \textbf{SRGCL}\footnote{\url{https://github.com/NTUDDSNLab/SRGCL}} (Self-Reinforced Graph Contrastive Learning), which leverages the encoder itself to \emph{detect} and \emph{select} high-quality positives:

\begin{itemize}
    \item \emph{Self-reinforced positive selection.} SRGCL forms a feedback loop in which high-quality positives iteratively refine the encoder, and the strengthened encoder, in turn, more accurately judges pair quality.
    \item \emph{Unified Positive Pair Generator (UPPG).} Multiple augmentation strategies are unified to produce diverse candidate pairs, from which SRGCL retains only those consistent with the encoder’s manifold-based similarity criterion.
    \item \emph{Manifold-aware, probabilistic optimization.} We embed the \emph{manifold hypothesis} into a probabilistic selector that gradually sharpens its acceptance distribution as representation power improves, eliminating false positives while keeping informative diversity.
    \item \emph{Plug-and-play design and empirical gains.} SRGCL integrates seamlessly into existing GCL pipelines and achieves consistent improvements on multiple graph-level benchmarks over the state of the art, without compromising generality across domains.
\end{itemize}

Overall, SRGCL demonstrates that rigorously filtering positive pairs is vital for contrastive objectives and offers a principled, self-reinforcing solution to realize this goal.

The remainder of the paper is organized as follows: Section~\ref{sec:background} introduces the terminology, reviews the fundamentals of GCL, and categorizes existing positive pair construction methods. Section~\ref{sec:method} presents the proposed SRGCL framework and its core components. In Section~\ref{sec:evaluation}, we conduct extensive experiments and analyses comparing SRGCL with state-of-the-art GCL methods. Finally, Section~\ref{sec:conclusion} concludes the paper and outlines directions for future work.

\section{Background and Related Works}
\label{sec:background}

\subsection{Terminologies and Notations}
\label{sec:terminologies}

We introduce the primary symbols and notations used consistently in this paper. Matrices are denoted by bold uppercase letters (e.g., \(\mathbf{X}\)), vectors by bold lowercase letters (e.g., \(\mathbf{v}\)), scalars by normal letters (e.g., \(m\), \(n\)).
Sets are represented by calligraphic letters (e.g., \(\mathcal{V}\)). 
We primarily examine undirected attributed graphs. We represent a graph as \(\mathcal{G} = \{\mathcal{V}, \mathcal{E}\}\), where \(\mathcal{V}\) is the node set \(\{v_i\}_{i=1}^n\) and \(\mathcal{E}\) is the edge set \(\{e_i\}_{i=1}^m\). Alternatively, in matrix form, we denote the graph as \(\mathcal{G} = (\mathbf{A}, \mathbf{X})\), where \(\mathbf{A} \in \mathbb{R}^{n \times n}\) is the adjacency matrix and \(\mathbf{X} \in \mathbb{R}^{n \times k}\) is the node feature matrix, with \(n\) as the number of nodes, \(m\) as the number of edges, and \(d\) as the feature dimension. 
When dealing with multiple graphs (e.g., in graph-level tasks), subscripts are used to differentiate between graphs and their corresponding components. For instance, \(p_i\) represents the positive pair of the \(i\)-th graph \(\mathcal{G}_i\).

\subsection{Graph Contrastive Learning}
\label{sec:gcl}

Graph Contrastive Learning (GCL) is a self-supervised framework designed to learn robust representations of graph-structured data by contrasting positive and negative sample pairs. Inspired by the success of contrastive learning in computer vision~\cite{he2020momentum, chen2020simple}, GCL derives supervisory signals directly from the data without relying on explicit labels.

To encode structural information, GCL employs Graph Neural Networks (GNNs), where a typical layer updates the representation of node \(v\) by aggregating information from its neighbors:
\begin{equation}
\mathbf{x}_v^{(l)} = \sigma\!\Big(\mathbf{W}^{(l)} \cdot \mathrm{AGG}\big(\{\mathbf{x}_{u}^{(l-1)} : u \in \mathcal{N}(v)\}\big)\Big),
\end{equation}
where \(\mathbf{x}_v^{(l)}\) is the node embedding at layer \(l\), \(\mathrm{AGG}(\cdot)\) is an aggregation function (e.g., mean or sum), \(\sigma\) is a nonlinear activation, and \(\mathbf{W}^{(l)}\) is a learnable weight matrix. A readout function, often an MLP, pools node embeddings to obtain a graph-level representation \(\mathbf{z}\).

Positive pairs are generated by applying different augmentations to the same node or graph, while negative pairs consist of different nodes or graphs. GCL typically adopts the InfoNCE loss~\cite{oord2018representation} to align positive pairs and separate negative ones:
\begin{equation}
\mathcal{L} = -\sum_i \log \frac{\exp(\mathrm{sim}(\mathbf{z}_i, \mathbf{z}_i^{+})/\tau)}{\sum_{j}\mathbb{I}_{j \neq i}\exp(\mathrm{sim}(\mathbf{z}_i, \mathbf{z}_j)/\tau)},
\end{equation}
where \(\mathrm{sim}(\cdot,\cdot)\) denotes a similarity measure (e.g., cosine), and \(\tau\) is a temperature parameter.

Positive pair quality is critical, as it provides the alignment signal necessary for learning perturbation-invariant and semantically meaningful embeddings. In the following sections, we discuss modern strategies for constructing high-quality positive pairs in GCL.

\begin{figure*}[t]
    \centering
    \includegraphics[width=0.9\linewidth]{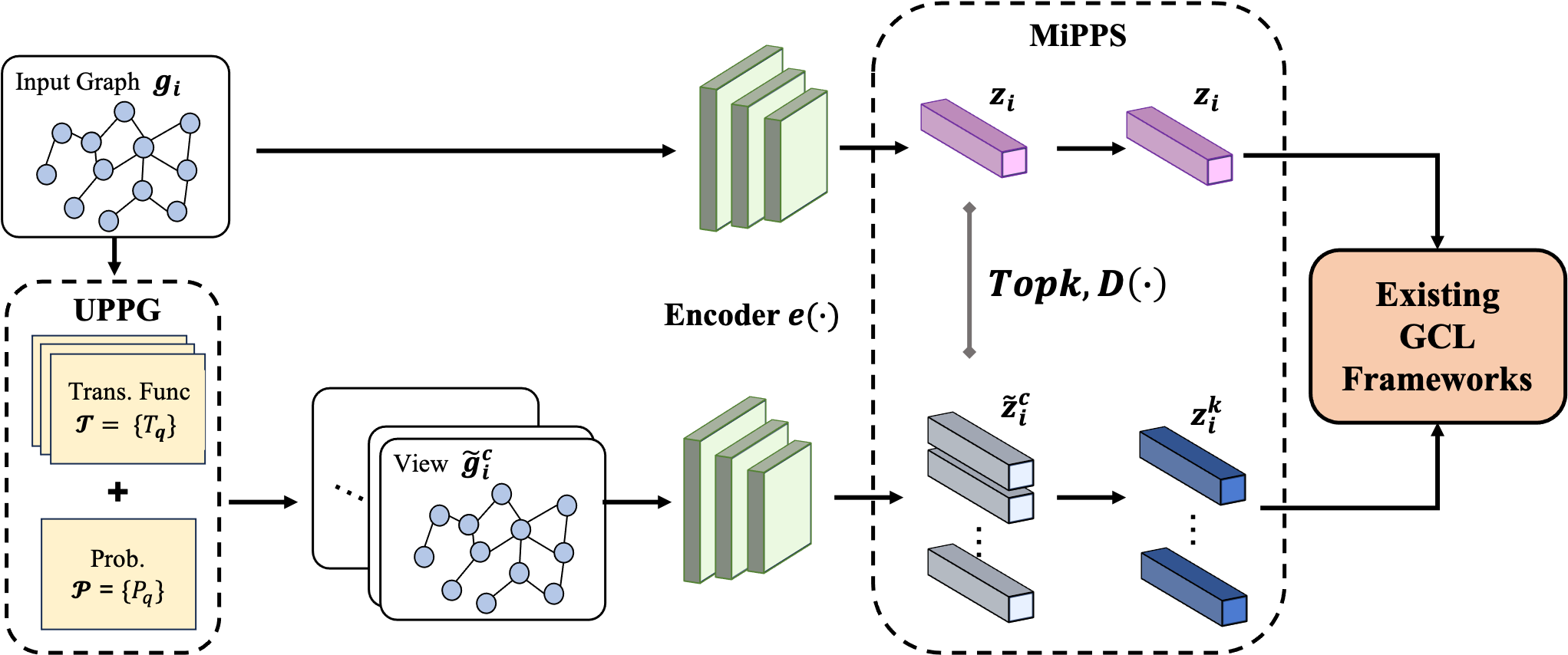}
    \caption{The proposed framework of SRGCL, which is the plug-in module for existing GCL frameworks. SRGCL comprises: (1) UPPG for generating positive pair candidate set $\mathcal{C}_{i}$, whose size is $c$, (2) GNN shared encoder $e(\cdot)$ for encoding the input graph and candidate positive pairs to vector $z_i, \tilde{z}_{i}^{c}$, and (3) MiPPS for selecting the final positive set $\mathcal{P}_{i}$ from $\mathcal{C}_{i}$.}
    \label{fig:architecture}
\end{figure*}

\subsection{Positive Pair Construction}
\label{sec:positive_pair_construction}

Positive Pair Construction (PPC) is a critical component of GCL, as it defines which instances are aligned in the representation space. In contrast to Negative Pair Construction (NPC), which typically treats other nodes or graphs as negatives, constructing high-quality positive pairs without label supervision remains a major challenge. Effective PPC is essential for learning meaningful and robust representations~\cite{hassani2020contrastive, yin2022autogcl, you2020graph, xia2022simgrace}. Positive pairs are generally derived from semantically or structurally similar graph instances through augmentations that preserve key properties. In this section, we categorize existing PPC strategies into \textit{rule-based} and \textit{learnable-based} approaches.

\textbf{Rule-based PPCs} rely on predefined augmentations such as node dropping, edge perturbation, feature masking, and diffusion-based transformations. GraphCL~\cite{you2020graph} introduces random node/edge dropping and subgraph sampling. Given the data-dependent performance of such augmentations, subsequent works propose more structured approaches. MVGRL~\cite{hassani2020contrastive} contrasts representations from adjacency and diffusion views to capture both local and global structures. InfoGraph~\cite{sun2019infograph} maximizes mutual information between global graph embeddings and substructures such as nodes, edges, and motifs. SimGRACE~\cite{xia2022simgrace} perturbs encoder weights directly, eliminating the need for graph-level augmentations and preserving semantic integrity.

\textbf{Learnable-based PPCs} automatically generate positive pairs via optimization-based strategies. JOAO~\cite{you2021graph} employs bi-level optimization to adaptively select augmentations and integrates an augmentation-aware projection head. JOAOv2~\cite{you2021graph} enhances this with an expanded augmentation pool and probabilistic selection. AD-GCL~\cite{suresh2021adversarial} learns edge-dropping probabilities through adversarial training, minimizing redundancy while preserving task-relevant features. AutoGCL~\cite{yin2022autogcl} introduces learnable view generators that produce semantically consistent augmented subgraphs, jointly trained with the encoder and classifier to optimize both structural variability and semantic alignment.

\section{Methodologies}
\label{sec:method}

To address the inconsistent quality of different PPC strategies among different datasets, 
we propose self-reinforced graph contrastive learning (SRGCL) framework, shown in Figure \ref{fig:architecture}.
The core idea of SRGCL is to generate and determine high-quality positive pairs with the GCL's own model.
During the training, 
we firstly generate multiple positive pairs through the Unified Positive Pair Generator (UPPG), 
which provides diverse PPC options.
After encoding these views to low-dimension feature vectors, 
the model selects the high-quality pair with Manifold-inspired Positive Pair Selector (MiPPS),
which leverages manifold hypothesis for determining high-quality positive pairs. 

The self-reinforcement comes from one intuition.
We believe that \emph{the high-quality positive pair helps the GCL model to encode graphs into an appropriate latent space. As the model becomes representative, the ability of model to determine high-quality positive pair will also increase}, 
which forms a virtuous cycle for training GCL model. (Figure \ref{fig:main_idea})

\subsection{Unified Positive Pair Generator}
\label{sec:UPPG}
 
The Unified Positive Pair Generator (UPPG) is designed to systematically consolidate various positive pair construction strategies in graph contrastive learning, encompassing both rule-based (e.g., GraphCL, MVGRL) and learning-based methods (e.g., AutoGCL, AD-GCL). For each graph \(g_i\), UPPG generates multiple perturbed views \(\{\tilde{g}_i^{1}, \tilde{g}_i^{2}, \ldots\}\) by invoking a set of transformation methods \(\mathcal{T} = \{T_q\}_{q=1}^{Q}\). Here, each \(T_q\) represents a distinct view generator, which can be either rule-based (e.g., random node dropping, edge perturbation) or learning-based (e.g., parameterized augmentations). UPPG then aggregates all candidate views into a set \(C_{i} = \{\tilde{g}_i^{1}, \tilde{g}_i^{2}, \ldots\}\), referred to as the \textbf{positive pair candidate set} for \(g_i\).

To regulate the diversity of the augmented views, UPPG incorporates a discrete probability distribution \(\mathbf{P} = \{p_1, p_2, \ldots, p_Q\}\), where each \(p_q\) corresponds to the probability of selecting the \(k\)-th transformation method \(T_q\). Consequently, for each graph \(g_i\), UPPG draws \(c\) samples (i.e., outputs \(c\) distinct views) according to \(\mathbf{P}\). Concretely, let \(\kappa_j\) be a random variable denoting the index of the transformation method selected for the \(j\)-th view of \(g_i\). We draw \(\kappa_j\) from the categorical distribution \(\mathbf{P}\), which can be expressed as:
\begin{equation}
\Pr(\kappa_j = q) \;=\; p_q, 
\quad \sum_{q=1}^{Q} p_q \;=\; 1.
\end{equation}
The resulting view \(\tilde{g}_i^{j}\) is obtained by applying the chosen transformation \(T_{Q_j}\) to \(g_i\):
\begin{equation}
\tilde{g}_i^{j} \;=\; T_{Q_j}\left(g_i\right), 
\quad j \in \{1,2,\ldots,c\}.
\end{equation}
Hence, the positive pair candidate set for the graph \(g_i\) is:
\begin{equation}
\mathcal{C}_i 
\;=\;
\left\{
\tilde{g}_i^{1}, \tilde{g}_i^{2}, \ldots, \tilde{g}_i^{c}
\right\}.
\end{equation}
Here, \(c\) denotes the total number of output views required by the contrastive learning framework. This probabilistic strategy enables UPPG to draw from the full spectrum of existing augmentation methods, thereby facilitating a comprehensive exploration of feature space perturbations. By unifying rule-based and learning-based transformations within a single pipeline, UPPG ensures that the training process can seamlessly leverage both expert-designed heuristics and adaptive augmentation mechanisms, ultimately yielding more robust and generalizable graph encoders.

\begin{table*}[ht]
    \centering

    \resizebox{\textwidth}{!}{%
    \begin{threeparttable}
        \caption{Performance Comparison with State-of-the-art Frameworks on the Graph Classification Task}

        \label{tab:comparison}
        \begin{tabular}{l|cccc|cccc}
        \midrule
        \midrule
        Methods & NCI1 & PROTEINS & DD & MUTAG & COLLAB & RDT-B & RDT-M5K & IMDB-B \\
        \midrule
        \midrule
        (No. Graphs) & 4,110 & 1,113 & 1,178 & 188 & 5,000 & 2,000 & 4,999 & 1,000 \\
        (No. Classes) & 2 & 2 & 2 & 2 & 3 & 2 & 5 & 2 \\
        (No. Node Dim) & 37 & 3 & 89 & 7 & 0 & 0 & 0 & 0 \\
        (Avg. Graph Size) & 29.87 & 39.06 & 284.32 & 14.29 & 74.49 & 429.63 & 508.52 & 19.77 \\
        \midrule
        \midrule
        GraphSAGE & 77.70 $\pm$ 1.50 & 75.90 $\pm$ 3.20 & 72.30 $\pm$ 0.00 & 85.10 $\pm$ 7.60 & 73.90 $\pm$ 1.70 & 84.30 $\pm$ 1.90 & 50.00 $\pm$ 1.30 & 68.80 $\pm$ 4.50 \\
        GCN & 80.20 $\pm$ 2.00 & 76.00 $\pm$ 3.20 & 74.00 $\pm$ 0.00 & 85.60 $\pm$ 5.80 & 79.00 $\pm$ 1.80 & 50.00 $\pm$ 0.00 & 20.00 $\pm$ 0.00 & - \\
        GIN-$0$ & 82.70 $\pm$ 1.70 & 76.20 $\pm$ 2.80 & 75.10 $\pm$ 0.00 & 89.40 $\pm$ 5.60 & 80.20 $\pm$ 1.90 & 92.40 $\pm$ 2.50 & 57.50 $\pm$ 1.50 & - \\
        GAT & - & - & - & 89.40 $\pm$ 6.10 & - & 85.20 $\pm$ 3.30 & - & 70.50 $\pm$ 2.30 \\
        \midrule
        GL & - & - & - & 81.66 $\pm$ 2.11 & - & 77.34 $\pm$ 0.18 & 41.01 $\pm$ 1.17 & 65.87 $\pm$ 0.98 \\
        WL & 80.01 $\pm$ 0.50 & 72.92 $\pm$ 0.56 & - & 80.72 $\pm$ 3.00 & - & 68.82 $\pm$ 0.41 & 46.06 $\pm$ 0.21 & 72.30 $\pm$ 3.44 \\
        DGK & 80.31 $\pm$ 0.46 & - & 73.50 $\pm$ 1.25 & 87.44 $\pm$ 2.72 & - & 78.04 $\pm$ 0.39 & 41.27 $\pm$ 0.18 & 66.96 $\pm$ 0.56 \\
        node2vec & 54.89 $\pm$ 1.61 & 57.49 $\pm$ 3.57 & - & 72.63 $\pm$ 10.2 & - & 71.48 $\pm$ 0.41 & 36.68 $\pm$ 0.42 & 55.26 $\pm$ 1.54 \\
        sub2vec & 52.84 $\pm$ 1.47 & 53.03 $\pm$ 5.55 & - & 61.05 $\pm$ 15.8 & - & 71.48 $\pm$ 0.32 & 36.68 $\pm$ 0.41 & 55.26 $\pm$ 1.54 \\
        graph2vec & 73.22 $\pm$ 1.81 & 73.3 $\pm$ 0.05 & - & 83.15 $\pm$ 9.25 & - & 75.78 $\pm$ 1.03 & 47.86 $\pm$ 0.26 & - \\
        \midrule
        InfoGraph & 76.20 $\pm$ 1.06 & 74.44 $\pm$ 0.31 & 72.85 $\pm$ 1.78 & 89.01 $\pm$ 1.13 & 70.65 $\pm$ 1.13 & 82.50 $\pm$ 1.42 & 53.46 $\pm$ 1.03 & 73.03 $\pm$ 0.87 \\
        JOAOv2 & 78.36 $\pm$ 0.53 & 74.07 $\pm$ 1.10 & 77.40 $\pm$ 1.15 & 87.67 $\pm$ 0.79 & 69.33 $\pm$ 0.34 & 86.42 $\pm$ 1.45 & 56.03 $\pm$ 0.27 & 70.83 $\pm$ 0.25 \\
        MVGRL & - & - & - & \underline{89.70 $\pm$ 1.10} & - & 84.50 $\pm$ 0.60 & - & 74.20 $\pm$ 0.70 \\
        SimGRACE & 79.12 $\pm$ 0.44 & 75.35 $\pm$ 0.79 & 77.44 $\pm$ 1.11 & 89.01 $\pm$ 1.31 & 71.72 $\pm$ 0.82 & 89.51 $\pm$ 0.89 & 55.91 $\pm$ 0.34 & 71.30 $\pm$ 0.77 \\
        ADGCL & 69.67 $\pm$ 0.51 & 73.59 $\pm$ 0.65 & 74.49 $\pm$ 0.52 & 89.25 $\pm$ 1.45 & 73.32 $\pm$ 0.61 & 85.52 $\pm$ 0.79 & 53.00 $\pm$ 0.82 & 71.57 $\pm$ 1.01 \\
        \midrule
        GraphCL & 77.87 $\pm$ 0.41 & 74.39 $\pm$ 0.45 & 78.62 $\pm$ 0.40 & 86.80 $\pm$ 1.34 & 71.36 $\pm$ 1.15 & 89.34 $\pm$ 0.53 & 55.99 $\pm$ 0.28 & 71.14 $\pm$ 0.44 \\
        GraphCL\_SR  & \textbf{80.97 $\pm$ 0.05} & 74.01 $\pm$ 0.44 & 77.93 $\pm$ 0.85 & \textbf{88.78 $\pm$ 1.08} & \underline{\textbf{73.36 $\pm$ 0.82}} & \underline{\textbf{90.05 $\pm$ 0.23}} & \underline{\textbf{56.36 $\pm$ 0.21}} & \textbf{72.69 $\pm$ 1.23} \\ 
        GraphCL\_SR (p) & \textbf{78.35 $\pm$ 0.26} & 73.42 $\pm$ 0.45 & \underline{\textbf{79.14 $\pm$ 0.63}} & \textbf{89.38 $\pm$ 0.59} & \textbf{71.4 $\pm$ 0.36} & \textbf{89.74 $\pm$ 0.05} & 55.41 $\pm$ 0.13 & \textbf{74.11 $\pm$ 0.33} \\
        \midrule
        AutoGCL & \underline{82.00 $\pm$ 0.29} & 75.80 $\pm$ 0.36 & 77.57 $\pm$ 0.60 & 88.64 $\pm$ 1.08 & 70.12 $\pm$ 0.68 & 88.58 $\pm$ 1.49 & 56.75 $\pm$ 0.18 & 73.30 $\pm$ 0.40 \\
        AutoGCL\_SR  & 81.62 $\pm$ 0.37 & 75.83 $\pm$ 0.30 & 77.47 $\pm$ 0.63 & \textbf{89.12 $\pm$ 0.96} & 68.20 $\pm$ 3.18 & 84.86 $\pm$ 5.06 & 52.64 $\pm$ 9.37 & 73.16 $\pm$ 0.66 \\ 
        AutoGCL\_SR (p) & 81.15 $\pm$ 0.29 & \underline{\textbf{75.91 $\pm$ 0.48}} & \textbf{77.76 $\pm$ 0.52} & \textbf{88.78 $\pm$ 0.92} & \textbf{70.68 $\pm$ 1.85} & \textbf{89.43 $\pm$ 1.42} & 53.68 $\pm$ 2.72 & \underline{\textbf{74.46 $\pm$ 0.37}} \\
        \bottomrule
        \end{tabular}%

        \begin{tablenotes}
        \item The first three rows illustrate the properties of each datasets. The rows from GraphSAGE to GAT are the supervised learning results, while the rest of the table is unsupervised learning results. The \textbf{bold} font represents that the model outperforms the original one with SRGCL plug-in; the \underline{underline} font shows the best performance among existing unsupervised GCL frameworks; "-" represents the result is missing or unavailable in the original paper.
        \end{tablenotes}
    \end{threeparttable}
    }
\end{table*}

\subsection{Manifold-inspired Positive Pair Selector}
\label{sec:manifold_hypothesis}

The manifold hypothesis \cite{tenenbaum2000global, bengio2013representation} is a cornerstone in modern machine learning and data science, asserting that real-world high-dimensional data often reside on or near a low-dimensional manifold embedded in the ambient space.  
For graph-level tasks, prior studies \cite{rubin2020manifold, wang2024manifold} have shown that graph datasets similarly exhibit manifold structures.  
Formally, let \(\mathcal{G} = \{ g_1, g_2, \ldots, g_m \} \subset \mathcal{X} \in \mathbb{R}^n\) be a set of graph data embedded in a manifold \(\mathcal{M} \subset \mathbb{R}^n\) of dimension \(d \ll n\).  
We assume there exist smooth functions \(f\) and \(e\) such that  
\begin{equation}
    f: \mathcal{Z} \rightarrow \mathcal{X}, \quad x = f(z),
    \quad\quad
    e: \mathcal{X} \rightarrow \mathcal{Z}, \quad z = e(x),
\end{equation}
where \(\mathcal{Z} \subset \mathbb{R}^d\) is the low-dimensional latent space, \(x \in \mathcal{X}\) and \(z \in \mathcal{Z}\).  By construction, any point \(x \in \mathcal{M}\) can be represented as \(x = f(z)\) for some \(z \in \mathcal{Z}\), thereby characterizing the manifold \(\mathcal{M}\) itself as:  
\begin{equation}
    \mathcal{M}
    \;=\;
    \bigl\{ x \in \mathcal{X} \,\big|\,
    \exists z \in \mathcal{Z}, \; x = f(z) \bigr\}.
\end{equation}

Under this geometric perspective, \emph{manifold smoothness} stipulates that two data points close in \(\mathcal{Z}\) are more likely to have similar labels than those lying far apart \cite{tenenbaum2000global, bengio2013representation}.  
Let \(\mathcal{C}_i\) denote the positive pair candidate set for a given graph \(g_i\), as described in Section~\ref{sec:UPPG}.
To identify high-quality positives, we first map each candidate \(\tilde{g}_i^{j}\) and the input graph \(g_i\) into \(\mathcal{Z}\) via the encoder \(e(\cdot)\).  
We then compute their Euclidean distance in the latent space, i.e. \(D\bigl(e(\tilde{g}_i^{j}), e(g_i)\bigr)\).  
Because of the manifold hypothesis, candidates with smaller distances in \(\mathcal{Z}\) are more likely to share the same class as \(g_i\).  

Hence, we define a \emph{top}-\(k\) selector that returns the \(k\) closest views from \(\mathcal{C}_i\).  
The selection procedure for the \textbf{positive pair set} \(\mathcal{P}_i\) can be formulated as:  
\begin{equation}
    \mathcal{P}_i 
    \;=\;
    \Bigl\{
    \underset{p_{i,j} \,\in\, \mathcal{C}_i}{\mathrm{top-}k}
    D\bigl(e(\tilde{g}_i^{j}),\, e(g_i)\bigr)
    \Bigr\},
    \quad
    1 \le i \le m, \quad 1 \le j \le c,
\end{equation}
where the operation \(\mathrm{top-}k\) returns the \(k\) smallest distances in \(\{D(e(\tilde{g}_i^{j}), e(g_i))\}_{j=1}^{c}\).  
Such a filtering step ensures that only the most \emph{manifold-consistent} candidates, i.e., those lying close to \(g_i\) in the low-dimensional space, enter the contrastive loss computation.  

Two theoretical perspectives reinforce this design.  
First, contrastive objectives (e.g., InfoNCE) induce representations that cluster similar data points in latent space, while pushing dissimilar points apart.  
Second, the smoothness property of the manifold hypothesis implies that small perturbations in \(\mathcal{X}\) or \(\mathcal{Z}\) lead to small changes in the label, thereby favoring close neighbors to share the same class.  
Concretely, if \(x_2 = x_1 + \nabla x\) is a small perturbation of \(x_1\), for a labeling function \(h(\cdot)\) we have:
\begin{equation}
    h\bigl(e(x_2)\bigr)
    \;\approx\;
    h\bigl(e(x_1)\bigr)
    \;+\;
    \nabla\! \bigl(h(e(x_1))\bigr)^{\mathsf{T}} \,
    J_{e}(x_1)\,\nabla x,
\end{equation}
where \(J_{e}(x_1)\) is the Jacobian of the encoder \(e\) at \(x_1\).  
If \(x_1\) and \(x_2\) lie close on the manifold, then \(\|\nabla x\|\) is small, implying \(h\bigl(e(x_1)\bigr)\) and \(h\bigl(e(x_2)\bigr)\) are nearly identical.  
By leveraging this property, the proposed top-\(k\) selection directly exploits manifold smoothness to ensure that only views conforming to the intrinsic geometry of the data are retained as positive pairs.

\subsection{Probabilistic Optimization on Selector}
\label{sec:probabilistic}

Although the manifold-inspired positive pair select indeed increases the GCL's accuracy by selecting better positive pairs, 
there are still room for improvement.
First, we observe that as the GCL model trains, its ability to encode graphs into an appropriate latent space strengthens. This suggests that early in training, the model struggles to properly position graphs in the latent space.  
Second, while the manifold hypothesis suggests that positive pairs closer to the input graph are more likely to share the same class, it does not guarantee that the closest positive pair always belongs to the same class. To address this, we introduce randomness into the positive pair selection process.  
SRGCL employs temperature decay (TD) \cite{warhaft1978experimental} to gradually reduce randomness as model confidence increases. This technique, widely used in applications such as nuclear physics, stellar modeling \cite{warhaft1978experimental}, and simulated annealing \cite{van1987simulated}, controls exploration ratios effectively. The probability \( P_{i,j}^{(t)} \) of selecting a candidate \( p_{i,j} \) as a positive pair in epoch \( t \) follows a Boltzmann-type (softmax) distribution \cite{harten1983upstream}, as shown below:

\begin{equation}
P_{i,j}^{(t)} = \frac{w_{i,j}(t)}{\sum_{\ell=1}^N w_{i,\ell}(t)} = \frac{\exp(-\frac{D(z_{i,j}, z_i)}{T(t)})}{\sum_{\ell=1}^N \exp(-\frac{D(z_{i,\ell}, z_i)}{T(t)})}
    \label{eqn:selector_td}
\end{equation}

\noindent
, where $w_{i,j}(t)$ is the unnormalized weight function of each $p_{i,j}$ at the iteration $t$;  $z_{i,j} = e(p_{i,j})$ is the low-dimension vector of candidate $p_{i,j}$.
$T(t) = T_0 e^{-st}$ is the geometric (exponential) decay function, where $e$ is the natural number; $s$ is the decay constant to determine the decay rate
,and $T_0$ is the initial temperature.
Once the probability distribution $\left\{ P_{i,j}^{(t)}\right\}$ is calculated, 
SRGCL samples $k$ (the number of selected positive pairs in an epoch) times from it without replacement to avoid duplicates.

\begin{figure*}[t]
    \centering
    \begin{subfigure}[b]{\linewidth}
        \includegraphics[width=\linewidth]{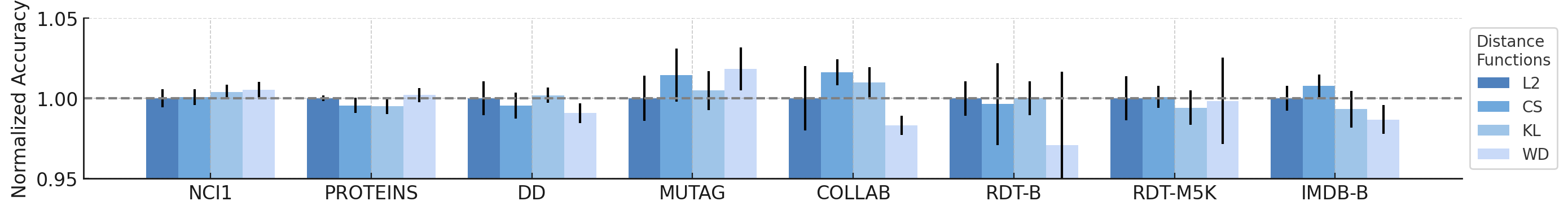}
        \caption{}
        \label{fig:dist_graphcl}
    \end{subfigure}
    \begin{subfigure}[b]{\linewidth}
        \includegraphics[width=\linewidth]{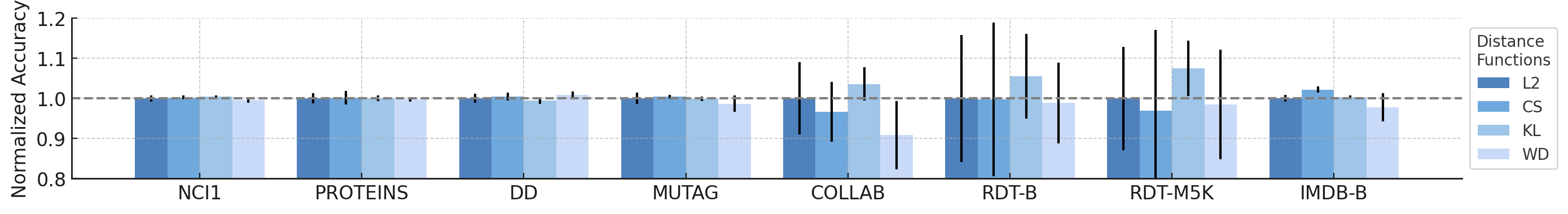}
        \caption{}
        \label{fig:dist_autogcl}
    \end{subfigure}
    \caption{Performance versus different distance functions across different datasets on Graph\_SR (a) AutoGCL\_SR (b). The accuracy is normalized to L2-norm configuration. The black vertical line is the standard deviation of each configuration.}
    \label{fig:distance_functions}
\end{figure*}

\section{SRGCL as an instance of EM Algorithm}
\label{sec:em_algorithm}

In this section, we show that the SRGCL can be understood from the perspective of an expectation-maximization (EM) algorithm. 
We will first introduce the concept of EM algorithm briefly.
We then introduce how we frame SRGCL as an instance of EM algorithm.

\subsection{General Expectation-Maximiation Algorithm}

The EM algorithm is a general-purpose framework for maximum-likelihood estimation in probabilistic models with latent variables.  
Given observed data \( \mathcal{X}=\{x_i\}_{i=1}^{N} \) and unobserved variables \( \mathcal{Z}=\{z_i\}_{i=1}^{N} \), EM iteratively maximizes the marginal log-likelihood
\begin{equation}
\ell(\theta)
  =\log p(\mathcal{X}\mid\theta)
  =\log\sum_{\mathcal{Z}} p(\mathcal{X},\mathcal{Z}\mid\theta)
\end{equation}
by alternating between two coupled optimizations.  

\textbf{E-step.}  
Compute the posterior \emph{responsibility} of the latent variables under the current parameters,  
\( q^{(t)}(\mathcal{Z}) = p(\mathcal{Z}\mid\mathcal{X},\theta^{(t)}) \), thereby forming the expected complete-data log-likelihood (the \(Q\)-function):

\begin{equation}
Q(\theta\mid\theta^{(t)})
   = \mathbb{E}_{q^{(t)}}\!\bigl[\log p(\mathcal{X},\mathcal{Z}\mid\theta)\bigr].
\end{equation}

\textbf{M-step.}  
Update the parameters by maximizing the surrogate objective:
\begin{equation}
\theta^{(t+1)}
   = \arg\max_{\theta}\; Q(\theta\mid\theta^{(t)})
\end{equation}

The objective/ model is typically updated via gradient‐based optimisation.  
For large data sets, evaluating the
full gradient across all observations can be computationally prohibitive.
\emph{Stochastic Expectation–Maximization} (SEM) \cite{balakrishnan2017statistical, papandreou2015weakly, zhu2017high} alleviates this
bottleneck by estimating the gradient on a small mini-batch of data, thereby
turning the M-step into a stochastic gradient ascent update while preserving,
in expectation, the monotonic improvement of the data likelihood.

\subsection{Framing SRGCL to SEM Algorithm}

\textbf{E-step: } 
Let $g_i$ denote the $i$‑th anchor graph and $g_{ij}'$ its $j$‑th candidate view $(j=1,\dots,C_i)$.
Introduce a binary latent variable $z_{ij}$, where

\begin{equation}
z_{ij} = \begin{cases} 1, & \text{if } g_{ij}' \text{ and } g_i \text{ belong to the same class},\\ 0, & \text{otherwise}. \end{cases}
\end{equation}

We denote the parameters of the GNN encoder as $\theta$; hence, 
the Q‑function for the SEM algorithm is $$ Q(\theta \mid \theta^{(t)}) = \mathbb{E} \bigl[ \log p\bigl(z_{ij}=1 \mid g_i,g_{ij}',\theta\bigr) \bigr]. $$

Because exact class labels are unavailable, we invoke the manifold‑smoothness assumption mentioned in last section \ref{sec:manifold_hypothesis}: candidate views that are closer in the learned representation space are more likely to share a class, 
which means the probability of these views to be true-positive is approximate to the distance between views and anchor graph. 
The probability $p$ can then be derived as: 
\begin{equation}
p\bigl(z_{ij}=1 \mid g_i,g_{ij}',\theta\bigr) \;\propto\; D\bigl(f_\theta(g_i),,f_\theta(g_{ij}')\bigr)
\end{equation}

,where $f_\theta(\cdot)$ is the encoder and $D(\cdot,\cdot)$ a suitable distance (Euclidean distance). 
Following common practice in large‑scale contrastive learning, we convert the soft posterior $\gamma_{ij}^{(t)} \equiv p(z_{ij}=1 \mid g_i,g_{ij}',\theta^{(t)})$ into a hard assignment that retains only the $k$ nearest neighbours:

\begin{equation}
S_i = \arg \text{top-k} \ D\bigl(f_{\theta^{(t)}}(g_i),,f_{\theta^{(t)}}(g_{ij}')\bigr)
\end{equation}

\begin{equation}
\gamma_{ij}^{'(t)} = \begin{cases} \dfrac{1}{k}, & j \in S_i^{},\\ 0, & j \notin S_i^{}. \end{cases}
\end{equation}

Thus each anchor is associated with exactly $k$ presumed positives, each weighted uniformly by $1/k$.

\textbf{M-step: }
Substituting $\gamma_{ij}'^{(t)}$ into the expected complete‑data log‑likelihood yields the surrogate objective

\begin{equation}
\begin{split}
\theta^{(t+1)} =\; \arg\max_{\theta} \sum_{i=1}^{|\mathcal V|} \sum_{j=1}^{C_i} \gamma_{ij}'^{(t)}, \mathcal L\bigl(f_\theta(g_i),f_\theta(g_{ij}')\bigr) \\
=\; \arg\max_{\theta} \sum_{i=1}^{|\mathcal V|} \frac{1}{k} \sum_{j\in S_i^{(t)}} \mathcal L\bigl(f_\theta(g_i),f_\theta(g_{ij}')\bigr)
\end{split}
\end{equation}

where $\mathcal{L}$ is any contrastive loss (e.g., InfoNCE) and $|\mathcal{V}|$ is the number of anchor graphs.
Consequently, the M‑step reduces to standard contrastive learning over the top‑$k$ closest candidate views.

By alternating the E‑step (which refines the latent assignments $z_{ij}$) with the M‑step (which updates $\theta$), the procedure progressively tightens the alignment of true positives in the embedding space, while pushing false positives away. Under mild regularity conditions, the sequence ${\theta^{(t)}}$ converges to a stationary point of the SEM objective.

\section{Experimental Results}
\label{sec:evaluation}

\subsection{Benchmarks}
\label{sec:benchmarks}

We evaluate our method on eight widely used graph datasets from the TUDataset collection~\cite{morris2020tudataset}, covering both biochemical (BI) and social network (SN) domains. BI datasets (NCI1, PROTEINS, DD, MUTAG) model molecular and protein structures, where nodes represent atoms or amino acids and edges indicate chemical bonds or spatial proximity. SN datasets (COLLAB, RDT-B, RDT-M5K, IMDB-B) capture social and collaboration networks, with nodes denoting individuals (e.g., researchers, users, actors) and edges reflecting co-authorship or interactions. These datasets span binary and multi-class classification tasks across diverse real-world scenarios. Detailed statistics are summarized in Table~\ref{tab:comparison}.

\subsection{Experiment Setup}
\label{sec:eval_protocol}


All experiments are conducted using PyTorch~\cite{paszke2019pytorch} and PyTorch Geometric~\cite{fey2019fast}. Since SRGCL serves as a plug-in module for existing graph contrastive learning (GCL) frameworks, we adopt the original training configurations (e.g., learning rate, batch size, optimizer) provided in their open-source implementations. SRGCL is evaluated on GraphCL and AutoGCL—GraphCL introduces four standard augmentations, while AutoGCL achieves state-of-the-art performance among GCL methods.

Each baseline is extended by integrating it with SRGCL. We denote the variant without probabilistic optimization as \texttt{Baseline\_SR}, and the version with probabilistic optimization as \texttt{Baseline\_SR(p)}. For GraphCL, we apply three augmentations—node dropping, edge perturbation, and attribute masking—within SRGCL. Subgraph augmentation is excluded due to its consistently large induced distance from the original graph, which diminishes its effectiveness for contrastive alignment in SRGCL.

In AutoGCL, the MiPPS module is inserted after the two view generators. Each generator independently produces \(c\) candidate views, which are then processed by MiPPS.

\textbf{SRGCL Hyperparameters.} The number of candidate positives \(c\) is set to 50, and the top-\(k\) selection parameter is set to 2. The temperature constant \(s\) is 0.4 for \texttt{GraphCL\_SR} and 0.8 for \texttt{AutoGCL\_SR}. MiPPS uses Euclidean distance as the similarity metric \(D(\cdot)\). When multiple transformations are available, the UPPG probability distribution \(P\) is set uniformly.

\textbf{Evaluation Protocol.} We follow the standard protocol from prior works~\cite{niepert2016learning, sun2019infograph, you2020graph, hassani2020contrastive}, using 10-fold cross-validation to report classification accuracy. Each experiment is repeated five times, and we report the mean accuracy and standard deviation. Models are trained for 30 epochs, and evaluated via a downstream support vector classifier (SVC)~\cite{brereton2010support} every 10 epochs, consistent with previous practice~\cite{you2020graph, hassani2020contrastive}.

\subsection{Comparisons with State-of-the-art Methods}
\label{sec:comparisons_SOTAs}

To evaluate the graph classification task, 
we compare our results with three graph kernels: Graphlet kernel (GL) \cite{shervashidze2009efficient}, Weisfeiler-Lehman sub-tree kernel (WL) \cite{shervashidze2011weisfeiler}, and deep graph kernels (DGK) \cite{yanardag2015deep}. 
We also compare our results with four classic supervised GNN models: GCN \cite{kipf2016semi}, GraphSAGE \cite{hamilton2017inductive}, GIN \cite{xu2018powerful}, and GAT \cite{velickovic2017graph}. 
In the unsupervised learning field, 
we categorized different methods into two types: traditional and GCL-based graph embedding algorithms. 
The traditional approaches comprise node2vec \cite{grover2016node2vec}, sub2vec \cite{adhikari2018sub2vec}, and graph2vec \cite{narayanan2017graph2vec}, 
while the GCL-based approaches can be further categorized into two types according to the positive pair construction: rule-based and learn-based GCL.
The rule-based GCLs comprise GraphCL \cite{you2020graph}, JOAOv2 \cite{you2021graph}, MVGRL \cite{hassani2020contrastive}, SimGrace \cite{xia2022simgrace}, and InfoGraph \cite{sun2019infograph}.
The learn-based GCLs include AD-GCL \cite{suresh2021adversarial} and AutoGCL \cite{yin2022autogcl}.
Besides the original paper of each framework, we also collected the experiment result from other survey papers \cite{errica2019fair, hassani2020contrastive}.
The comparison is shown in Table \ref{tab:comparison}.

Empirical results demonstrate that SRGCL consistently enhances the performance of its base models, both with and without probabilistic optimization. Specifically, GraphCL\_SR improves performance on six out of eight datasets and achieves state-of-the-art (SOTA) results on three of them. Furthermore, incorporating probabilistic optimization in GraphCL\_SR(p) leads to additional performance gains on three datasets; however, it results in performance degradation on four others.  

Conversely, while AutoGCL\_SR surpasses AutoGCL on only one out of eight datasets, AutoGCL\_SR(p) exhibits significant improvements, outperforming AutoGCL on six datasets. This observation suggests a strong affinity between AutoGCL and probabilistic optimization, indicating that AutoGCL benefits more substantially from this optimization technique.

We noticed that the performance of AutoGCL on RDT-B and RDT-M5K, including the original model, has remarkable fluctuation. 
Based on an analysis of AutoGCL's open-source code \cite{autogcl_github}, we found that the significant oscillations observed in large-scale graphs such as RDT-B and RDT-M5K stem from the fact that its generated views retain only 20\% of the original graph’s size. Compared to smaller graphs, the removal of 80\% of the nodes has a more substantial impact on larger graphs, increasing the likelihood of information loss, which in turn leads to pronounced fluctuations.

\begin{figure}[t]
    \centering
    \begin{subfigure}[b]{\linewidth}
        \includegraphics[width=\linewidth]{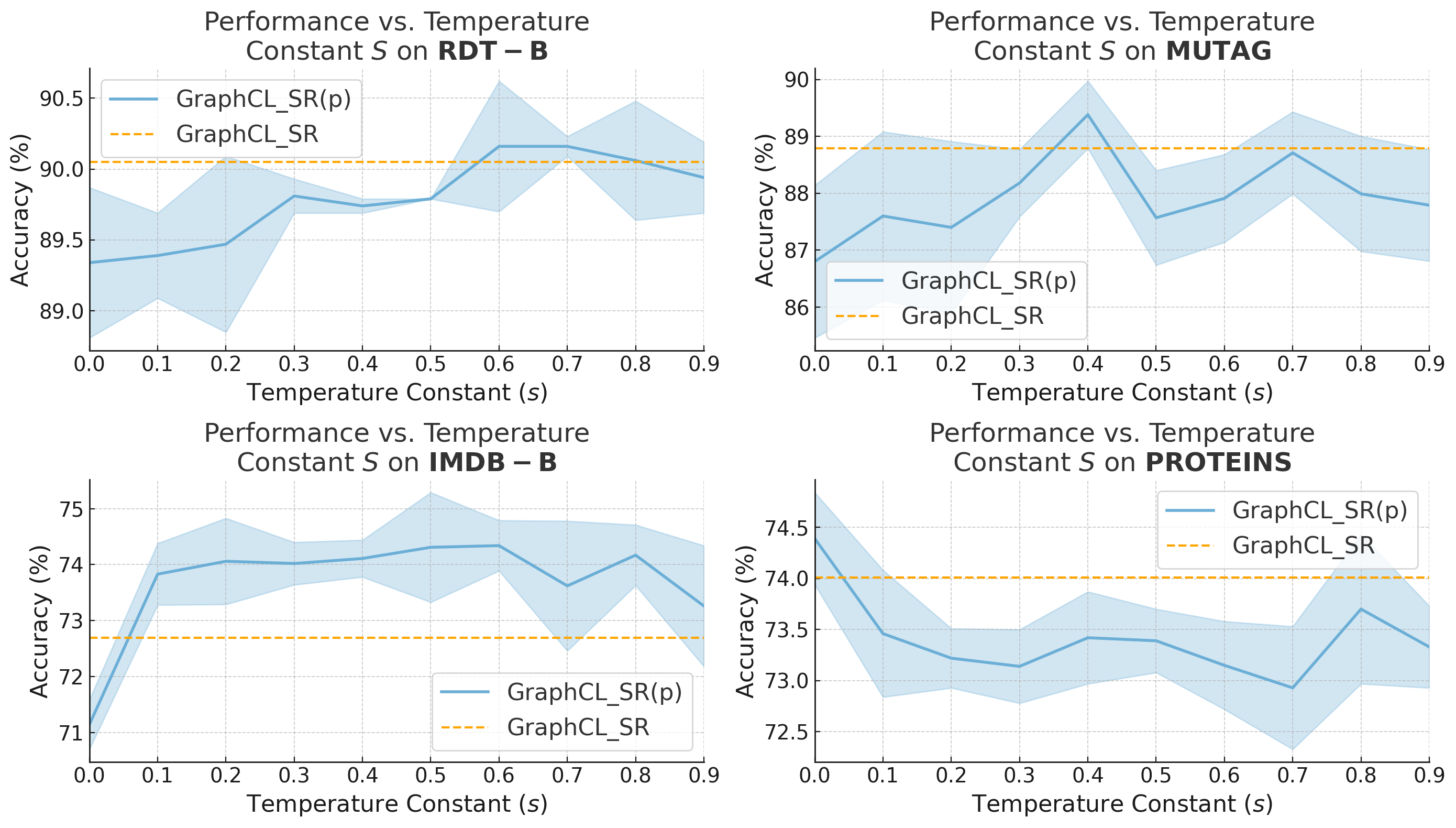}
        \caption{}
        \label{fig:temperature_constant_graphcl}
    \end{subfigure}
    \begin{subfigure}[b]{\linewidth}
        \includegraphics[width=\linewidth]{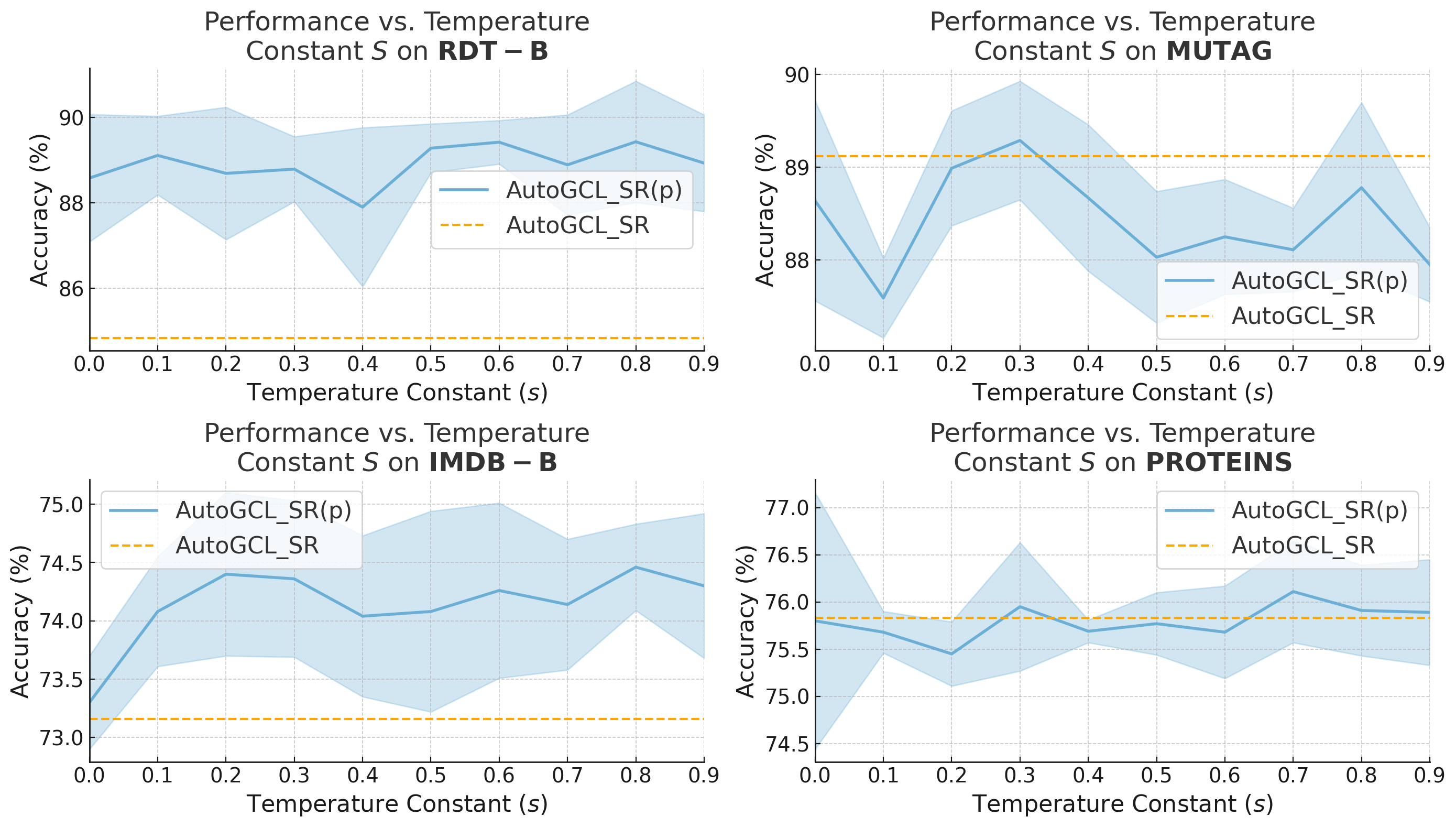}
        \caption{}
        \label{fig:temperature_constant_autogcl}
    \end{subfigure}
    
    \caption{Performance versus temperature constant $s$ on GraphCL\_SR (a) and AutoGCL\_SR (b). The light blue tunnel part is the standard deviation of SRGCL run in 5 times. The orange dashed line is the average accuracy of SRGCL without probabilistic optimization. Once $s=0$, the SRGCL degrades to its original design, which can be considered as the fully random implementation. }
    \label{fig:temperature_constant}
\end{figure}

\subsection{Ablation Studies}
\label{sec:ablation}

\begin{figure}
    \centering
    \includegraphics[width=\linewidth]{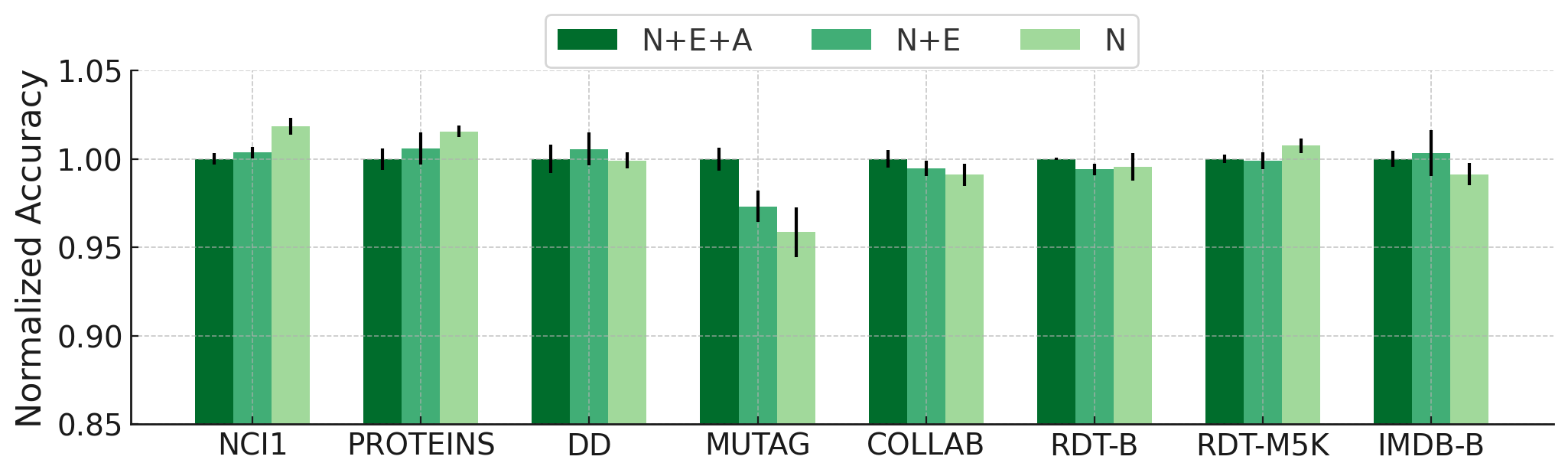}
    \caption{Performance versus augmentation combination across 8 datasets on GraphCL\_SR (p). The accuracy is normalized to N+E+A. The black vertical line is the standard deviation of each configuration.}
    \label{fig:diff_aug}
\end{figure}

\begin{figure}
    \centering
    \begin{subfigure}[b]{\linewidth}
        \includegraphics[width=\linewidth]{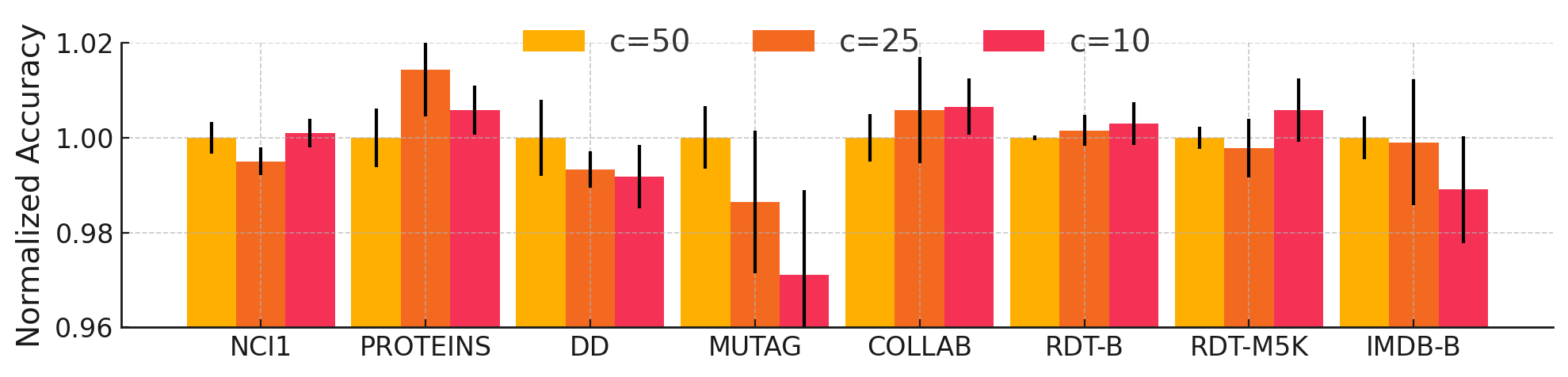}
        \caption{}
        \label{fig:diffc_graphcl}
    \end{subfigure}
    \begin{subfigure}[b]{\linewidth}
        \includegraphics[width=\linewidth]{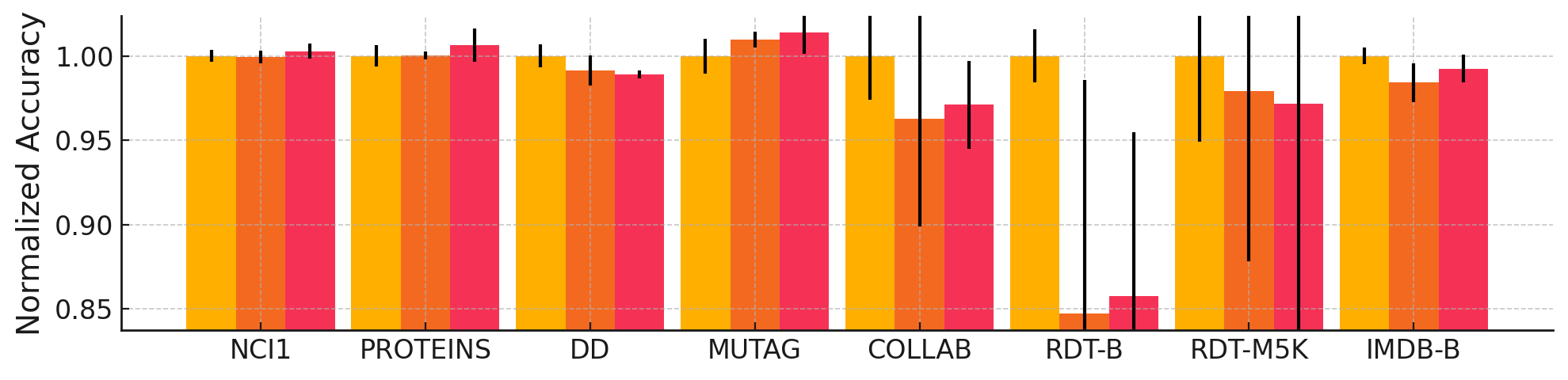}
        \caption{}
        \label{fig:diffc_autogcl}
    \end{subfigure}
    \caption{Performance versus different candidate set size $c$ across 8 datasets on GraphCL\_SR (a) and AutoGCL\_SR (b). The accuracy is normalized to $c=50$. The black vertical line is the standard deviation of each configuration.}
    \label{fig:diffc}
\end{figure}

\subsubsection{Effect of Distance Function}

We evaluated four distance (similarity) functions, categorized into two types: distance-based and probability-based measures. For distance-based methods, we selected Euclidean distance (L2) and cosine similarity (CS). 
We considered KL Divergence (KL) \cite{csiszar1975divergence} and Wasserstein Distance (WD) \cite{vaserstein1969markov} for probability-based methods. 
These functions were assessed on SRGCL, with the results presented in Figure \ref{fig:distance_functions}.  
While different distance functions produce slight variations in performance, WD consistently yields the poorest results, particularly on SN-type datasets such as COLLAB, RDT-B, and IMDB-B. Furthermore, WD increases the instability of SRGCL on SN-type datasets, indicating its limited suitability for these tasks.

\subsubsection{Effect of Probabilistic Optimization}

We investigated the impact of different levels of randomization in the MiPPS module by adjusting the temperature constant \( s \) from 0 (fully random) to 0.9 in increments of 0.1. A higher value of \( s \) results in a rapid reduction in randomness as training progresses. To assess this effect, we conducted experiments using GraphCL\_SR and AutoGCL\_SR on four selected datasets: RDT-B, IMDB-B, MUTAG, and PROTEINS, which include two SN-type and two BI-type datasets. The results are presented in Figure \ref{fig:temperature_constant}.  

Overall, GraphCL\_SR, both with and without probabilistic optimization, consistently outperforms the original GraphCL across most datasets, particularly on RDT-B, IMDB-B, and MUTAG. Notably, the performance of GraphCL\_SR on RDT-B peaks at approximately 90.2\% when \( s=0.6 \), while IMDB-B achieves its highest accuracy of around 74.4\% at the same temperature setting. In the case of AutoGCL\_SR, the model demonstrates superior performance over AutoGCL on SN-type datasets (RDT-B and IMDB-B); however, its performance exhibits noticeable fluctuations on BI-type datasets.  

Regarding the selection of the temperature constant, SN-type datasets exhibit a preference for moderate values of \( s \), whereas BI-type datasets show only a slight sensitivity to this parameter. Probabilistic optimization notably enhances performance on RDT-B and IMDB-B in AutoGCL\_SR (p). In contrast, its effect on BI-type datasets is less stable, leading to performance fluctuations in both GraphCL\_SR (p) and AutoGCL\_SR (p). Specifically, MUTAG achieves peak performance at approximately 89.2\% and 89.3\% with GraphCL\_SR (p) at \( s=0.4 \) and AutoGCL\_SR (p) at \( s=0.3 \), respectively. Meanwhile, PROTEINS attains its highest accuracy at 73.7\% and 76.25\% with GraphCL\_SR (p) at \( s=0.8 \) and AutoGCL\_SR (p) at \( s=0.7 \), respectively. These variations in optimal \( s \) values across datasets highlight the necessity of dataset-specific hyperparameter tuning to maximize performance.

\subsubsection{Effect of Transformation Method}


We ablated the augmentation methods on Graph\_SR (p) to investigate the effect of different combination. 
We denote \textbf{N} as node dropping, \textbf{E} as edge perturbing, and \textbf{A} as attribute masking.
We developed three different variants : \textbf{N}, \textbf{N+E}, and \textbf{N+E+A}. 
The total candidate set size $c$ is set to 50 for all configurations. 
We generate equal amount of candidates to each augmentation if there are multiple augmentation integrated.  
The result is shown in Figure \ref{fig:diff_aug}.
Except for MUTAG, whose performance decreases through the augmentation methods ablate, 
the performance of different augmentation combinations have approximate performance.
The result also validates that MiPPS can select proper positive pair in almost every scenario.

\subsubsection{Effect of Candidate Positive Pair Set Size $c$}. 
We examine whether reducing the candidate set size affects the selector’s ability to identify high-quality positive pairs by decreasing \( c \) to 25 and 10, as shown in Figure \ref{fig:diffc}. A smaller \( c \) increases SRGCL's instability, reflected in higher accuracy variance, and adds computational pressure on SN-type datasets, particularly RDT-B and RDT-M5K in AutoGCL\_SR. AutoGCL\_SR's accuracy drops by up to 15\% on RDT-B when \( c = 25 \).  

GraphCL\_SR is generally less sensitive to \( c \) changes, except for a significant drop on MUTAG. While increasing candidate pairs improves SRGCL's stability, it also raises memory and computational costs, highlighting a trade-off between stability and efficiency.

\section{Overhead Estimation}
\label{sec:overhead}

\begin{figure}
    \centering
    \begin{subfigure}[b]{\linewidth}
        \includegraphics[width=\linewidth]{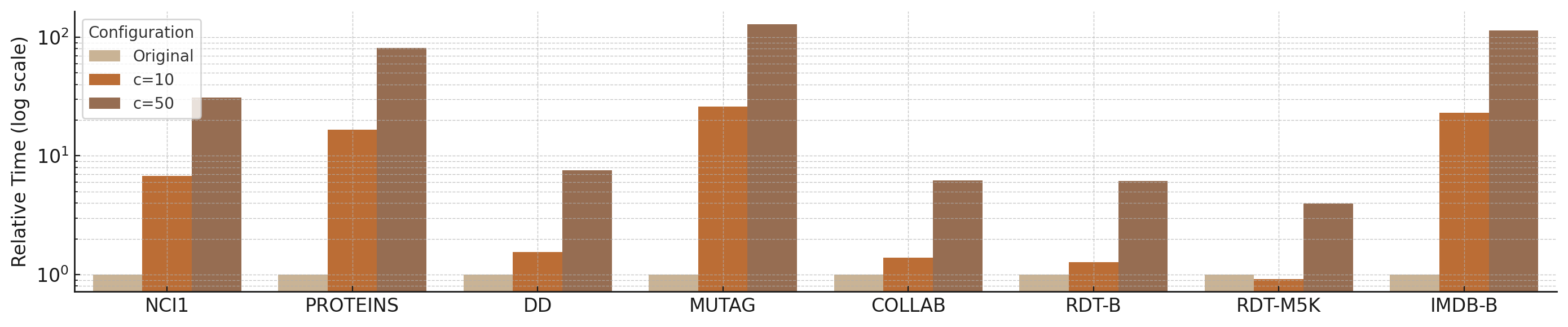}
        \caption{}
        \label{fig:overhead_time_graphcl}
    \end{subfigure}
    \begin{subfigure}[b]{\linewidth}
        \includegraphics[width=\linewidth]{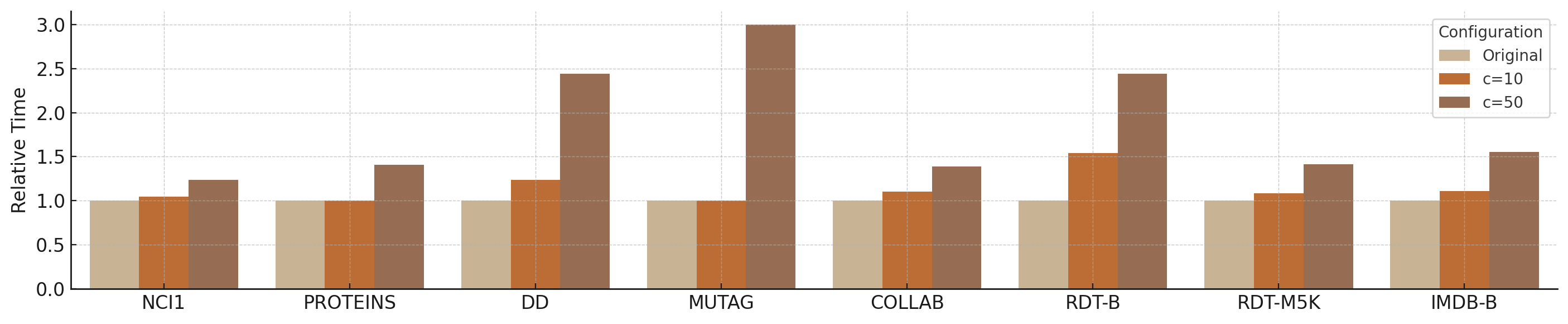}
        \caption{}
        \label{fig:overhead_time_autogcl}
    \end{subfigure}
    \caption{Normalized runtime of GraphCL (a) and AutoGCL 
 (b) with and without SRGCL on eight datasets. Results are shown for SRGCL with $c = 10$ and $c = 50$. GraphCL uses a log scale due to larger overhead, while AutoGCL shows minimal increase and is plotted on a linear scale.}
    \label{fig:overhead_time}
\end{figure}

\begin{table}[t]
\centering
\caption{Memory usage in GPU(in MB) per dataset for GraphCL and AutoGCL under different $c$ values}
\label{tab:memory_usage}
\resizebox{\columnwidth}{!}{%
\begin{tabular}{lccc|ccc}
\toprule
\textbf{Dataset} & \multicolumn{3}{c|}{\textbf{GraphCL}} & \multicolumn{3}{c}{\textbf{AutoGCL}} \\
\cmidrule(lr){2-4} \cmidrule(lr){5-7}
& \textbf{c=2} & \textbf{c=10} & \textbf{c=50} & \textbf{c=2} & \textbf{c=10} & \textbf{c=50} \\
\midrule
NCI1     & 510  & 510  & 510  & 406   & 868   & 1,630  \\
PROTEINS & 448  & 544  & 544  & 712   & 922   & 1,756  \\
DD       & 908  & 876  & 934  & 2,590  & 4,838  & 12,424 \\
MUTAG    & 498  & 500  & 500  & 610   & 692   & 1,066  \\
COLLAB   & 992  & 828  & 898  & 3,816  & 5,618  & 7,722  \\
RDT-B    & 914  & 944  & 952  & 4,506  & 7,126  & 22,828 \\
RDT-M5K  & 976  & 986  & 996  & 4,726  & 7,642  & 22,086 \\
IMDB-B   & 522  & 504  & 504  & 646   & 808   & 1,456  \\
\bottomrule
\end{tabular}%
}
\end{table}

Figure~\ref{fig:overhead_time} and Table~\ref{tab:memory_usage} present the runtime and memory overhead of integrating SRGCL into GraphCL and AutoGCL. The results indicate that SRGCL incurs significant runtime overhead on GraphCL, particularly at higher values of the hyperparameter $c$, due to host-side candidate generation and CPU-GPU data transfer becoming a bottleneck. In contrast, AutoGCL maintains a runtime overhead within 2× across all datasets, even at $c = 50$, highlighting its efficiency and robustness—likely attributed to its adaptive, device-oriented design.

Regarding memory usage, GraphCL exhibits minimal GPU memory overhead as augmented candidates are generated on the host. Conversely, AutoGCL performs candidate generation on the GPU, leading to increased memory consumption as $c$ grows. Overall, while SRGCL enhances contrastive learning, its integration is more practical and scalable with AutoGCL, especially in resource-constrained environments.

\section{Conclusion and Future Works}
\label{sec:conclusion}

We introduce SRGCL, a self-reinforced graph contrastive learning plug-in framework. The central premise of SRGCL is that high-quality positive pairs enhance the model's capacity, thereby improving its ability to discriminate the quality of positive pairs, ultimately forming a self-reinforcing cycle. Grounded in the manifold hypothesis, SRGCL leverages the model's learnable encoder to adaptively assess and select high-quality positive pairs.  

Empirical evaluations demonstrate that SRGCL significantly improves performance and even achieves state-of-the-art results on widely adopted graph classification benchmarks. However, our analysis indicates substantial room for further enhancement. Specifically, reducing memory and computational overhead, as well as optimizing application-specific hyperparameters, present promising directions for future research. Despite these limitations, SRGCL establishes the efficacy of high-quality positive pairs in enhancing GCL performance. Moreover, it introduces a novel paradigm for self-reinforced GCL frameworks, paving the way for a new avenue of research in graph contrastive learning.

\bibliographystyle{IEEEtran}
\bibliography{refs}

\begin{thebibliography}{10}
\providecommand{\url}[1]{#1}
\csname url@samestyle\endcsname
\providecommand{\newblock}{\relax}
\providecommand{\bibinfo}[2]{#2}
\providecommand{\BIBentrySTDinterwordspacing}{\spaceskip=0pt\relax}
\providecommand{\BIBentryALTinterwordstretchfactor}{4}
\providecommand{\BIBentryALTinterwordspacing}{\spaceskip=\fontdimen2\font plus
\BIBentryALTinterwordstretchfactor\fontdimen3\font minus \fontdimen4\font\relax}
\providecommand{\BIBforeignlanguage}[2]{{%
\expandafter\ifx\csname l@#1\endcsname\relax
\typeout{** WARNING: IEEEtran.bst: No hyphenation pattern has been}%
\typeout{** loaded for the language `#1'. Using the pattern for}%
\typeout{** the default language instead.}%
\else
\language=\csname l@#1\endcsname
\fi
#2}}
\providecommand{\BIBdecl}{\relax}
\BIBdecl

\bibitem{myers2014information}
S.~A. Myers, A.~Sharma, P.~Gupta, and J.~Lin, ``Information network or social network? the structure of the twitter follow graph,'' in \emph{Proceedings of the 23rd international conference on world wide web}, 2014, pp. 493--498.

\bibitem{borgwardt2005protein}
K.~M. Borgwardt, C.~S. Ong, S.~Sch{\"o}nauer, S.~Vishwanathan, A.~J. Smola, and H.-P. Kriegel, ``Protein function prediction via graph kernels,'' \emph{Bioinformatics}, vol.~21, no. suppl\_1, pp. i47--i56, 2005.

\bibitem{hogan2021knowledge}
A.~Hogan, E.~Blomqvist, M.~Cochez, C.~d’Amato, G.~D. Melo, C.~Gutierrez, S.~Kirrane, J.~E.~L. Gayo, R.~Navigli, S.~Neumaier \emph{et~al.}, ``Knowledge graphs,'' \emph{ACM Computing Surveys (Csur)}, vol.~54, no.~4, pp. 1--37, 2021.

\bibitem{fan2019graph}
W.~Fan, Y.~Ma, Q.~Li, Y.~He, E.~Zhao, J.~Tang, and D.~Yin, ``Graph neural networks for social recommendation,'' in \emph{The world wide web conference}, 2019, pp. 417--426.

\bibitem{you2020graph}
Y.~You, T.~Chen, Y.~Sui, T.~Chen, Z.~Wang, and Y.~Shen, ``Graph contrastive learning with augmentations,'' \emph{Advances in neural information processing systems}, vol.~33, pp. 5812--5823, 2020.

\bibitem{hassani2020contrastive}
K.~Hassani and A.~H. Khasahmadi, ``Contrastive multi-view representation learning on graphs,'' in \emph{International conference on machine learning}.\hskip 1em plus 0.5em minus 0.4em\relax PMLR, 2020, pp. 4116--4126.

\bibitem{sun2019infograph}
F.-Y. Sun, J.~Hoffmann, V.~Verma, and J.~Tang, ``Infograph: Unsupervised and semi-supervised graph-level representation learning via mutual information maximization,'' \emph{arXiv preprint arXiv:1908.01000}, 2019.

\bibitem{xia2022simgrace}
J.~Xia, L.~Wu, J.~Chen, B.~Hu, and S.~Z. Li, ``Simgrace: A simple framework for graph contrastive learning without data augmentation,'' in \emph{Proceedings of the ACM Web Conference 2022}, 2022, pp. 1070--1079.

\bibitem{you2021graph}
Y.~You, T.~Chen, Y.~Shen, and Z.~Wang, ``Graph contrastive learning automated,'' in \emph{International Conference on Machine Learning}.\hskip 1em plus 0.5em minus 0.4em\relax PMLR, 2021, pp. 12\,121--12\,132.

\bibitem{suresh2021adversarial}
S.~Suresh, P.~Li, C.~Hao, and J.~Neville, ``Adversarial graph augmentation to improve graph contrastive learning,'' \emph{Advances in Neural Information Processing Systems}, vol.~34, pp. 15\,920--15\,933, 2021.

\bibitem{yin2022autogcl}
Y.~Yin, Q.~Wang, S.~Huang, H.~Xiong, and X.~Zhang, ``Autogcl: Automated graph contrastive learning via learnable view generators,'' in \emph{Proceedings of the AAAI conference on artificial intelligence}, vol.~36, no.~8, 2022, pp. 8892--8900.

\bibitem{debnath1991structure}
A.~K. Debnath, R.~L. Lopez~de Compadre, G.~Debnath, A.~J. Shusterman, and C.~Hansch, ``Structure-activity relationship of mutagenic aromatic and heteroaromatic nitro compounds. correlation with molecular orbital energies and hydrophobicity,'' \emph{Journal of medicinal chemistry}, vol.~34, no.~2, pp. 786--797, 1991.

\bibitem{he2020momentum}
K.~He, H.~Fan, Y.~Wu, S.~Xie, and R.~Girshick, ``Momentum contrast for unsupervised visual representation learning,'' in \emph{Proceedings of the IEEE/CVF conference on computer vision and pattern recognition}, 2020, pp. 9729--9738.

\bibitem{chen2020simple}
T.~Chen, S.~Kornblith, M.~Norouzi, and G.~Hinton, ``A simple framework for contrastive learning of visual representations,'' in \emph{International conference on machine learning}.\hskip 1em plus 0.5em minus 0.4em\relax PMLR, 2020, pp. 1597--1607.

\bibitem{oord2018representation}
A.~v.~d. Oord, Y.~Li, and O.~Vinyals, ``Representation learning with contrastive predictive coding,'' \emph{arXiv preprint arXiv:1807.03748}, 2018.

\bibitem{tenenbaum2000global}
J.~B. Tenenbaum, V.~d. Silva, and J.~C. Langford, ``A global geometric framework for nonlinear dimensionality reduction,'' \emph{science}, vol. 290, no. 5500, pp. 2319--2323, 2000.

\bibitem{bengio2013representation}
Y.~Bengio, A.~Courville, and P.~Vincent, ``Representation learning: A review and new perspectives,'' \emph{IEEE transactions on pattern analysis and machine intelligence}, vol.~35, no.~8, pp. 1798--1828, 2013.

\bibitem{rubin2020manifold}
P.~Rubin-Delanchy, ``Manifold structure in graph embeddings,'' \emph{Advances in neural information processing systems}, vol.~33, pp. 11\,687--11\,699, 2020.

\bibitem{wang2024manifold}
Z.~Wang, J.~Cervino, and A.~Ribeiro, ``A manifold perspective on the statistical generalization of graph neural networks,'' \emph{arXiv preprint arXiv:2406.05225}, 2024.

\bibitem{warhaft1978experimental}
Z.~Warhaft and J.~Lumley, ``An experimental study of the decay of temperature fluctuations in grid-generated turbulence,'' \emph{Journal of Fluid Mechanics}, vol.~88, no.~4, pp. 659--684, 1978.

\bibitem{van1987simulated}
P.~J. Van~Laarhoven, E.~H. Aarts, P.~J. van Laarhoven, and E.~H. Aarts, \emph{Simulated annealing}.\hskip 1em plus 0.5em minus 0.4em\relax Springer, 1987.

\bibitem{harten1983upstream}
A.~Harten, P.~D. Lax, and B.~v. Leer, ``On upstream differencing and godunov-type schemes for hyperbolic conservation laws,'' \emph{SIAM review}, vol.~25, no.~1, pp. 35--61, 1983.

\bibitem{balakrishnan2017statistical}
S.~Balakrishnan, M.~J. Wainwright, and B.~Yu, ``Statistical guarantees for the em algorithm: From population to sample-based analysis,'' 2017.

\bibitem{papandreou2015weakly}
G.~Papandreou, L.-C. Chen, K.~P. Murphy, and A.~L. Yuille, ``Weakly-and semi-supervised learning of a deep convolutional network for semantic image segmentation,'' in \emph{Proceedings of the IEEE international conference on computer vision}, 2015, pp. 1742--1750.

\bibitem{zhu2017high}
R.~Zhu, L.~Wang, C.~Zhai, and Q.~Gu, ``High-dimensional variance-reduced stochastic gradient expectation-maximization algorithm,'' in \emph{International Conference on Machine Learning}.\hskip 1em plus 0.5em minus 0.4em\relax PMLR, 2017, pp. 4180--4188.

\bibitem{morris2020tudataset}
C.~Morris, N.~M. Kriege, F.~Bause, K.~Kersting, P.~Mutzel, and M.~Neumann, ``Tudataset: A collection of benchmark datasets for learning with graphs,'' \emph{arXiv preprint arXiv:2007.08663}, 2020.

\bibitem{paszke2019pytorch}
A.~Paszke, S.~Gross, F.~Massa, A.~Lerer, J.~Bradbury, G.~Chanan, T.~Killeen, Z.~Lin, N.~Gimelshein, L.~Antiga \emph{et~al.}, ``Pytorch: An imperative style, high-performance deep learning library,'' \emph{Advances in neural information processing systems}, vol.~32, 2019.

\bibitem{fey2019fast}
M.~Fey and J.~E. Lenssen, ``Fast graph representation learning with pytorch geometric,'' \emph{arXiv preprint arXiv:1903.02428}, 2019.

\bibitem{niepert2016learning}
M.~Niepert, M.~Ahmed, and K.~Kutzkov, ``Learning convolutional neural networks for graphs,'' in \emph{International conference on machine learning}.\hskip 1em plus 0.5em minus 0.4em\relax PMLR, 2016, pp. 2014--2023.

\bibitem{brereton2010support}
R.~G. Brereton and G.~R. Lloyd, ``Support vector machines for classification and regression,'' \emph{Analyst}, vol. 135, no.~2, pp. 230--267, 2010.

\bibitem{shervashidze2009efficient}
N.~Shervashidze, S.~Vishwanathan, T.~Petri, K.~Mehlhorn, and K.~Borgwardt, ``Efficient graphlet kernels for large graph comparison,'' in \emph{Artificial intelligence and statistics}.\hskip 1em plus 0.5em minus 0.4em\relax PMLR, 2009, pp. 488--495.

\bibitem{shervashidze2011weisfeiler}
N.~Shervashidze, P.~Schweitzer, E.~J. Van~Leeuwen, K.~Mehlhorn, and K.~M. Borgwardt, ``Weisfeiler-lehman graph kernels.'' \emph{Journal of Machine Learning Research}, vol.~12, no.~9, 2011.

\bibitem{yanardag2015deep}
P.~Yanardag and S.~Vishwanathan, ``Deep graph kernels,'' in \emph{Proceedings of the 21th ACM SIGKDD international conference on knowledge discovery and data mining}, 2015, pp. 1365--1374.

\bibitem{kipf2016semi}
T.~N. Kipf and M.~Welling, ``Semi-supervised classification with graph convolutional networks,'' \emph{arXiv preprint arXiv:1609.02907}, 2016.

\bibitem{hamilton2017inductive}
W.~Hamilton, Z.~Ying, and J.~Leskovec, ``Inductive representation learning on large graphs,'' \emph{Advances in neural information processing systems}, vol.~30, 2017.

\bibitem{xu2018powerful}
K.~Xu, W.~Hu, J.~Leskovec, and S.~Jegelka, ``How powerful are graph neural networks?'' \emph{arXiv preprint arXiv:1810.00826}, 2018.

\bibitem{velickovic2017graph}
P.~Velickovic, G.~Cucurull, A.~Casanova, A.~Romero, P.~Lio, Y.~Bengio \emph{et~al.}, ``Graph attention networks,'' \emph{stat}, vol. 1050, no.~20, pp. 10--48\,550, 2017.

\bibitem{grover2016node2vec}
A.~Grover and J.~Leskovec, ``node2vec: Scalable feature learning for networks,'' in \emph{Proceedings of the 22nd ACM SIGKDD international conference on Knowledge discovery and data mining}, 2016, pp. 855--864.

\bibitem{adhikari2018sub2vec}
B.~Adhikari, Y.~Zhang, N.~Ramakrishnan, and B.~A. Prakash, ``Sub2vec: Feature learning for subgraphs,'' in \emph{Advances in Knowledge Discovery and Data Mining: 22nd Pacific-Asia Conference, PAKDD 2018, Melbourne, VIC, Australia, June 3-6, 2018, Proceedings, Part II 22}.\hskip 1em plus 0.5em minus 0.4em\relax Springer, 2018, pp. 170--182.

\bibitem{narayanan2017graph2vec}
A.~Narayanan, M.~Chandramohan, R.~Venkatesan, L.~Chen, Y.~Liu, and S.~Jaiswal, ``graph2vec: Learning distributed representations of graphs,'' \emph{arXiv preprint arXiv:1707.05005}, 2017.

\bibitem{errica2019fair}
F.~Errica, M.~Podda, D.~Bacciu, and A.~Micheli, ``A fair comparison of graph neural networks for graph classification,'' \emph{arXiv preprint arXiv:1912.09893}, 2019.

\bibitem{autogcl_github}
\BIBentryALTinterwordspacing
{Yin, Yihang and Wang, Qingzhong and Huang, Siyu and Xiong, Haoyi and Zhang, Xiang}. Autogcl github repository. [Online]. Available: \url{https://github.com/Somedaywilldo/AutoGCL}
\BIBentrySTDinterwordspacing

\bibitem{csiszar1975divergence}
I.~Csisz{\'a}r, ``I-divergence geometry of probability distributions and minimization problems,'' \emph{The annals of probability}, pp. 146--158, 1975.

\bibitem{vaserstein1969markov}
L.~N. Vaserstein, ``Markov processes over denumerable products of spaces, describing large systems of automata,'' \emph{Problemy Peredachi Informatsii}, vol.~5, no.~3, pp. 64--72, 1969.

\end{thebibliography}

\end{document}